\documentclass[conference]{IEEEtran}
\usepackage{amsmath,amssymb,amsfonts}
\usepackage{algorithmic}
\usepackage{graphicx}
\usepackage{textcomp}
\usepackage{xcolor}
\usepackage[hyphens]{url}
\usepackage{xfrac}

\def\BibTeX{{\rm B\kern-.05em{\sc i\kern-.025em b}\kern-.08em
    T\kern-.1667em\lower.7ex\hbox{E}\kern-.125emX}}

\usepackage{amsthm}
\usepackage{bm}
\usepackage{multirow}
\usepackage[capitalise]{cleveref}
\usepackage[sort, numbers, compress]{natbib}
\usepackage[]{algorithm2e}
\usepackage{booktabs} 
\usepackage{subcaption}
\usepackage{placeins}

\title{
Breaking reCAPTCHAv2
}

\author{
\IEEEauthorblockN{Andreas Plesner}
\IEEEauthorblockA{
\textit{ETH Zurich}, Switzerland \\
aplesner@ethz.ch}
\and
\IEEEauthorblockN{Tobias Vontobel}
\IEEEauthorblockA{
\textit{ETH Zurich}, Switzerland \\
votobias@student.ethz.ch}
\and
\IEEEauthorblockN{Roger Wattenhofer}
\IEEEauthorblockA{
\textit{ETH Zurich}, Switzerland \\
wattenhofer@ethz.ch}
}

\begin{document}

\newcommand{\recaptcha}{\mbox{reCAPTCHAv2}}

\maketitle

\begin{abstract}
Our work examines the efficacy of employing advanced machine learning methods to solve captchas from Google's \recaptcha{} system. We evaluate the effectiveness of automated systems in solving captchas by utilizing advanced YOLO models for image segmentation and classification. Our main result is that we can solve 100\% of the captchas, while previous work only solved 68-71\%. Furthermore,
our findings suggest that there is no significant difference in the number of challenges humans and bots must solve to pass the captchas in \recaptcha{}. This implies that current AI technologies can exploit advanced image-based captchas. We also look under the hood of \recaptcha{}, and find evidence that \recaptcha{} is heavily based on cookie and browser history data when evaluating whether a user is human or not. 
The code is provided alongside this paper.\footnote{https://github.com/aplesner/Breaking-reCAPTCHAv2}\footnote{Corresponding author: Andreas Plesner}\footnote{Accepted at COMPSAC 2024}

\end{abstract}


\section{Introduction}

The challenge of distinguishing between humans and machines has become a critical aspect of online security. Captchas (``Completely Automated Public Turing Tests to Tell Computers and Humans Apart'') have emerged as the front-line defense against automated bots and malicious activities on the Internet. 

\begin{figure*}[ht!]
    \centering
    \begin{subfigure}[t]{0.27\linewidth}
        \includegraphics[width=\textwidth]{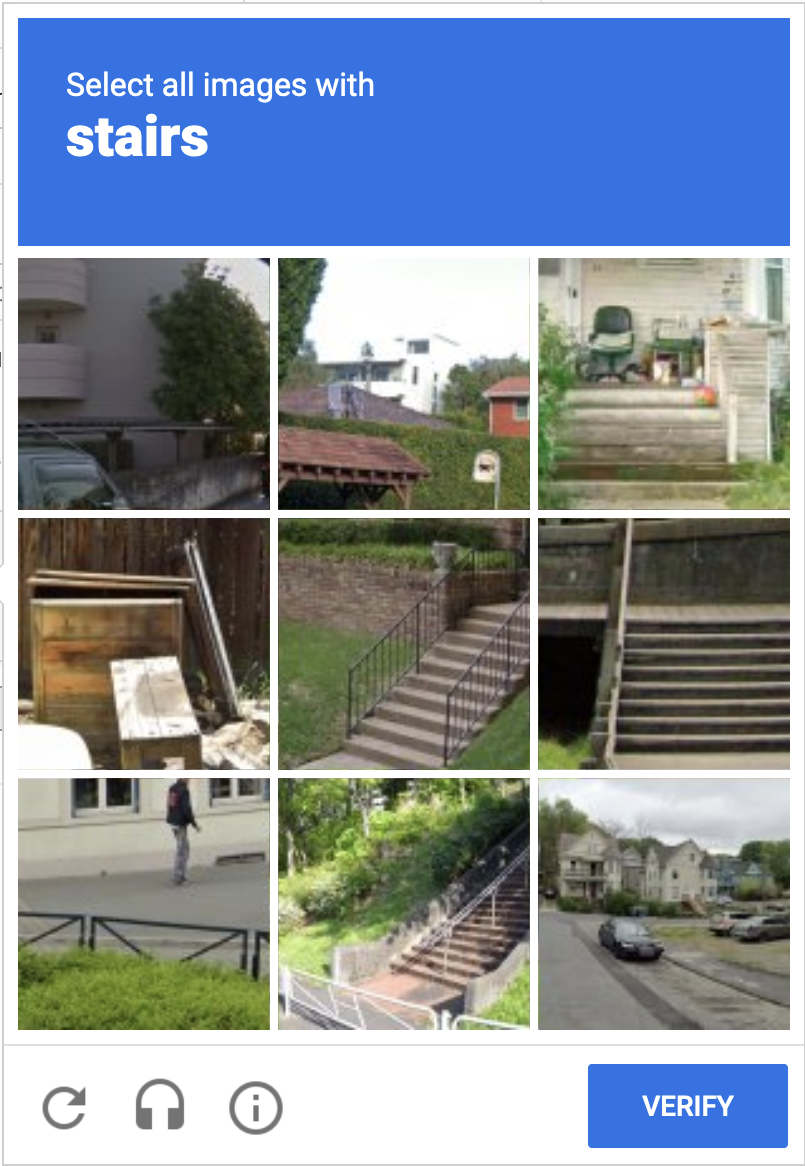}
        \caption{Type 1 captcha challenge example displaying a 3 by 3 grid of static images with target class "stairs". The user must identify all images containing stairs.}
        \label{fig:type1captcha}
    \end{subfigure}
    \hfill 
    \begin{subfigure}[t]{0.27\linewidth}
        \includegraphics[width=\textwidth]{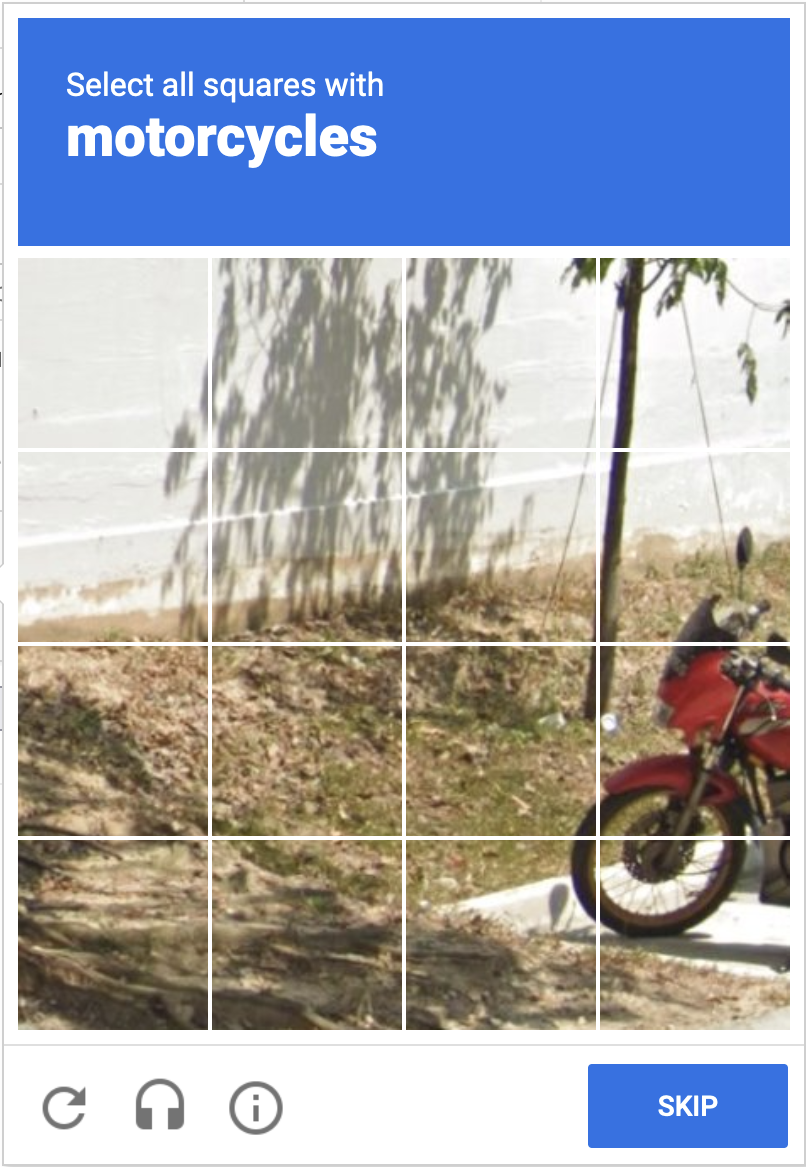}
        \caption{Type 2 captcha challenge example displaying a single image divided into a 4 by 4 grid with target class motorcycles. The user selects all squares containing motorcycles.}
        \label{fig:type2captcha}
    \end{subfigure}
    \hfill 
    \begin{subfigure}[t]{0.27\linewidth}
        \includegraphics[width=\textwidth]{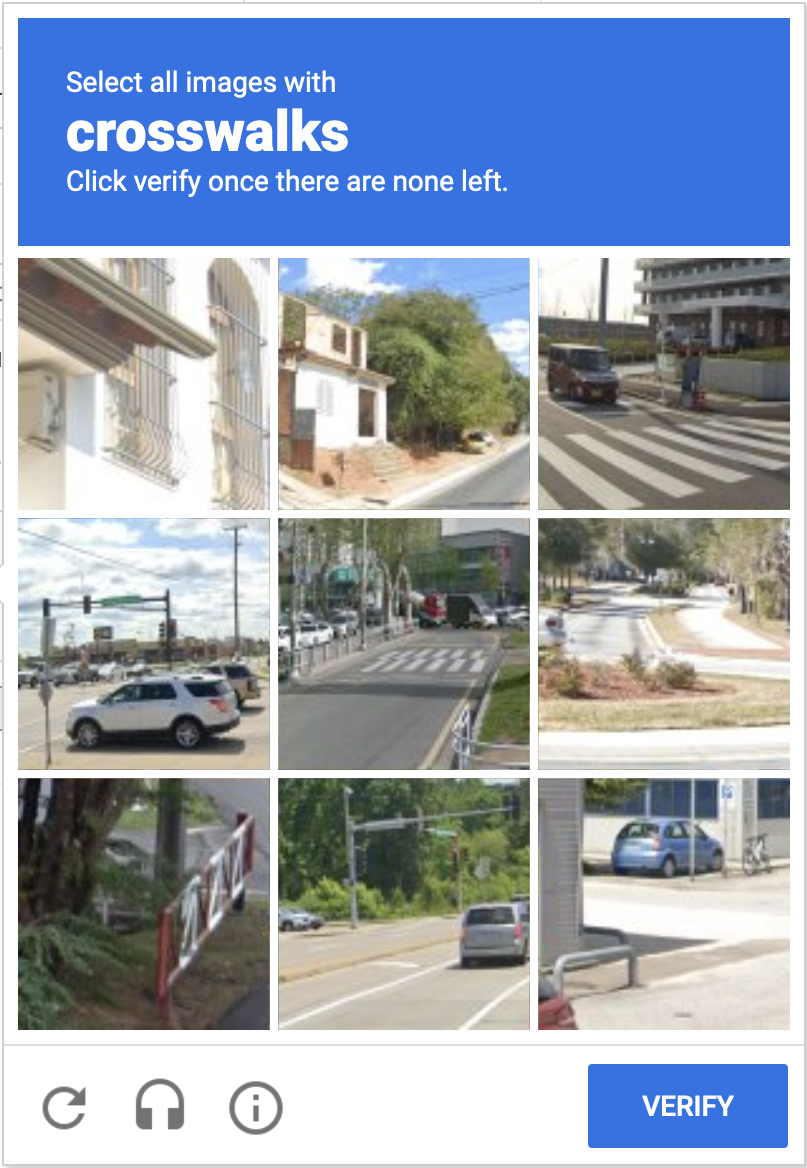}
        \caption{Type 3 captcha challenge example displaying a 3 by 3 grid of images with target class crosswalks; the images are replaced when the user clicks on them. The user must select all images with crosswalks.}
        \label{fig:type3captcha}
    \end{subfigure}
    \caption{Examples of the three different captcha type challenges used by Google's \recaptcha{}. Each type presents a unique challenge for users to solve to determine whether the user is a bot or not.}
    \label{fig:types_of_captcha}
\end{figure*}

In this work, we focus on Google's \recaptcha{} system \cite{recaptchav2,recaptchav3}, see also \cref{fig:types_of_captcha}. Our decision to use \recaptcha{} is based on its widespread use, indicating its significant role in protecting against automated threats. Moreover, \recaptcha{} is technically advanced, with an excellent trade-off between user experience and security, making it a preferred option for many websites. 
Because of this, Google's \recaptcha{} is a good representative of image-based captcha technology.

This project aims to analyze the effectiveness of Google's \recaptcha{} in rejecting bots using advanced deep learning models such as YOLO models.
Our main result is that we can solve 100\% of the captchas, while previous work solved 68-71\%. In addition, we find evidence that there is no significant difference in the number of challenges required by humans and bots to solve captchas in \recaptcha{}, if anything bots are better than humans. Our study also finds that \recaptcha{} is heavily based on cookie and browser history data when evaluating whether a user is human or not.

While automatically solving captchas sounds like a mundane job, there is also a deeper philosophical angle. In some sense, \textit{a good captcha marks the exact boundary between the most intelligent machine and the least intelligent human}. To be effective, literally every human above a certain age, independent of language and cultural background, must be able to solve the captcha. A captcha should never lock out humans. As machine learning models close in on human capabilities, finding good captchas has become more difficult. Bluntly, this paper shows that we are now officially in the age beyond captchas. 

In the past, captchas have often been (ab)used to have humans label data. This goes back to early text recognition captchas. Google's \recaptcha{} is suspected to be useful in training self-driving cars, as a large number of tests are traffic related. Dozens of companies are publicly working on developing self-driving cars, including Google with Waymo.
Waymo self-driving cars are already a reality. And they are on the verge of really tackling all possible traffic situations, with freeways currently being tested in Phoenix. Consequently, it is not surprising that \recaptcha{} can be solved. However, while it was always clear that Google can solve its captchas, we show that anybody can do so by just cleverly applying publicly available software. 

Google and others declared the end of captchas some years ago. Already \recaptcha{} usually only asks established users to check a box that they are ``not a robot.'' If enough evidence (browsing history, mouse movements, etc.) confirms that the user is indeed not a robot, \recaptcha{} will grant access directly. Only in a case where \recaptcha{} is insecure do the image recognition tests come into play. These image recognition tasks are quite dreaded by many users. A simple search finds plenty of testimony videos where unhappy users cannot solve a long series of \recaptcha{} tests. 

In fact, with Google's newest reCAPTCHAv3, a completely ``captcha-less captcha'' has already existed for several years. It decides the human vs. robot question purely on past interactions. However, some humans will ultimately not have sufficiently convincing browsing data and credentials. These unfortunate users may automatically and instantly be locked out of potentially vital internet services. In this case, to still gain access to a service, web pages often fall back to \recaptcha{} \cite{recaptchaFAQ, stackoverflowRecaptcha}. This is why \recaptcha{} still plays an important role in today's seemingly captcha-free world and deserves our attention.

\section{Related work}
\paragraph{Early Text-based Captchas}
Initially, captchas were text-based and designed primarily to prevent automated spam and abuse on websites. However, the introduction of powerful machine learning algorithms has made it possible to automate the solving of many text-based challenges. Some influential works in this area include the work of \citet{von2003captcha} and \citet{von2004telling} that laid the foundation for the development of captchas, and the work of \citet{chellapilla2004} and \cite{bursztein2014end} demonstrate the early success of machine learning in breaking text-based captchas. 

Researchers have used different machine learning algorithms, such as neural networks, generative adversarial networks (GANs), and convolutional neural networks (CNNs), to overcome text-based captchas \cite{WANG2021181, noury2020deepcaptcha}. These algorithms have achieved high success rates, leading to more advanced captcha schemes, such as image-based captchas being developed. The evolution of captcha problems, transitioning from text-based to image-based recognition, highlights the ongoing competition between security measures and solving algorithms.

\paragraph{Breaking reCAPTCHAv2}
The studies conducted by \citet{wang2020using,259735} and \citet{sivakorn2016m} are key references for this study, as they explore the weaknesses of Google's \recaptcha{} when faced with advanced deep learning models. These studies thoroughly examine the strength of image-based captchas, providing a detailed analysis of how the design of captchas interacts with the advanced capabilities of AI solvers. In these studies, the researchers were able to solve 68-71\% of captchas \cite{wang2020using, sivakorn2016m}.
The mentioned works thoroughly record the effectiveness of deep learning models in solving image-based captchas. The knowledge obtained from these influential works is crucial in determining the course of this research \cite{wang2020using,259735}.

Several open-source projects have aimed to automate solving Google's \recaptcha{} using machine learning techniques. The ''Recaptcha V2 Solver'' repository employs a combination of the BLIP language model for image captioning and YOLOv3 for object detection, achieving a success rate of 23-32\% on \recaptcha{} challenges \cite{artistrazh2023}. Similarly, the ''RecaptchaV2-IA-Solver'' repository utilizes the YOLOv8 model for object detection and supports both dynamic and one-time selection CAPTCHAs, but does not report specific success rates \cite{lunapy2023}. The ``GoodByeCaptcha'' library takes a different approach, using speech recognition APIs like Mozilla's DeepSpeech and Microsoft Azure's Speech-to-Text to solve audio reCAPTCHAs, while also incorporating image recognition for image-based challenges \cite{mackey2023}. In contrast, our work focuses solely on image-based reCAPTCHAv2 and achieves a 100\% success rate using a fine-tuned YOLOv8 model for image segmentation and classification tasks. We also conduct a more comprehensive analysis, comparing bot and human performance, examining the impact of browser cookies, and highlighting the importance of realistic mouse movements and VPN usage for evading detection.\footnote{We were unable to run the code in these libraries.}

Google has also released reCAPTCHAv3, which does not ask the user to solve any challenges. Instead, reCAPTCHAv3 analyzes how the user interacts with the website and uses this information to calculate a risk score. The back end of the website can then restrict the user if the score falls below a certain threshold. The key is that this is completely hidden from the user when interacting with the website \cite{GoogleRecaptchav2, GoogleRecaptchav3}.

\paragraph{Deep Learning for Image Classification}
\citet{NIPS2012_c399862d} introduced convolutional neural networks that drastically improved the state-of-the-art for image classification. Later, \citet{vaswani2017attention} introduced ResNet models, making models capable of high-accuracy image classification and object detection, setting a new state-of-the-art. Meanwhile, \citet{vaswani2017attention} introduced the Transformer architecture which has fueled many of the later advancements across domains such as vision \cite{khan2022transformers}, natural language \cite{brown2020language, openai2023gpt4} and image generation \cite{betker2023improving}. 

However, these models are computationally heavy, so this project uses the YOLO v8 model \cite{Jocher_Ultralytics_YOLO_2023}. YOLO v8 is an accurate and efficient model that belongs to the lineage of the 'You Only Look Once' models \cite{redmon2016look}. It is well known for its ability to detect objects in real-time \cite{shafiee2017fast}. This study specifically uses the YOLO v8 segmentation and classification models. The selection of YOLO v8 is in line with the objective of the study to evaluate image-based captcha security using the latest and most sophisticated machine learning technologies, guaranteeing a thorough and up-to-date examination. Moreover, YOLO models can be used on devices with limited computational power, allowing for large-scale attacks by malicious users.

\paragraph{Advancement in AI Models}
At the end of 2022, OpenAI released ChatGPT, an LLM which, along with later iterations, has demonstrated the ability of AI to solve tasks that involve abstract thinking \cite{brown2020language,openai2023gpt4,touvron2023llama}. Recently, DeepMind released AlphaGeometry, showcasing how LLMs could solve abstract geometry problems from the International Math Olympiad, IMO, on par with the IMO gold medalists \cite{trinh2024solving}. And \citet{radford2021learning} showed that computers can learn to represent images using natural language, while \citet{betker2023improving} presents a model capable of generating art. This shows how \textit{artificial} intelligence is rapidly catching up to human intelligence. 

\paragraph{Advancement in Captchas}
Aside from traditional text-based and image-based captchas, the domain also includes several innovative variations, such as Arkose Labs' FunCAPTCHA and MatchKey \cite{arkosematchkey}, AWS WAF captcha \cite{awswafcaptcha}, and hCaptcha \cite{hcaptcha}. Arkose Labs' FunCAPTCHA and MatchKey offer engaging minigames and advanced key-pattern analysis, respectively, providing unique user experiences. Integration of AWS WAF captcha with Web Application Firewalls Highlights its focus on robust security measures. hCaptcha, often used as an alternative to Google's reCAPTCHA, emphasizes user privacy and rewards for website owners. Each of these types of captcha shows different approaches to balancing user experience, security, and privacy, reflecting the diverse strategies in modern captcha design. \citet{5504799} highlights that as captchas have become more difficult for computers, the result is that many new captchas are difficult for humans.

\citet{osadchy2017no} created a series of captchas based on images specifically designed to be difficult for image classification models by generating adversarial examples \cite{kurakin2018adversarial}. The issue with this method is that it needs to be tuned to each classifier.

An area that shows promise in giving challenges that are suitable for determining whether a user is human or not is the Abstract Reasoning Challenge, ARC, \cite{ARC}. These challenges are difficult for computers to solve, with the best programs only solving 31\% while humans can solve 80\% of the challenges \cite{lab42arc}. 

\paragraph{Audio Captchas}
Audio captchas are essential to ensure accessibility for users with visual impairments, in addition to visual captcha challenges. An influential study in this field by \citet{NIPS2008_12092a75} devises a machine learning methodology to effectively solve audio captchas. They achieved significant success by successfully bypassing captchas with an accuracy rate of up to 71\% using techniques such as AdaBoost, SVM, and k-NN \cite{cortes1995support,fix1989discriminatory,freund1995desicion}. Consequently, there exists a restricted threshold for the number of audio captchas that a user can successfully solve in reCAPTCHA.

\section{Understanding reCAPTCHAv2}
This section presents the \recaptcha{} in detail together with the data used for this project.

\subsection{Background}
Google's \recaptcha{} consists of three different challenge types, each developed to test a specific aspect of visual reasoning. The type 1 captcha challenge, depicted in \cref{fig:type1captcha}, is a classification task that requires the user to determine whether each image on a static 3 by 3 grid contains the target object or not. On the other hand, the type 2 captcha challenge, as shown in \cref{fig:type2captcha}, is notable for being an image segmentation task. The challenge presents a single static image that is divided into a grid of 4 rows and 4 columns. Users are asked to divide and recognize specific parts of the image that are relevant to the challenge. Each type presents a unique task, either classification or segmentation, that requires specific approaches to automated solving. The type 3 captcha challenge, shown in \cref{fig:type3captcha}, is similar to type 1 in its grid layout, but incorporates dynamic images that refresh upon interaction, also requiring classification.

\subsection{Data}\label{sec:Data}
This subsection discusses the data used for the machine learning model.

\paragraph{Classification}
Both type 1 and type 3 captchas use image grids in their classification tasks, as seen in \cref{fig:type1captcha} and \cref{fig:type3captcha}, which are required for the identification of particular subjects. 11,774 labeled images from a publicly accessible dataset were used to fine-tune the machine learning model for this task \cite{mandourah_recaptcha_dataset}. Furthermore, due to the bot's operations, a dataset of the current captcha images was collected.

Using a pre-trained YOLOv8 for classification, a semi-automated tool was created to speed up the labeling process. To provide training data of the highest quality, this model first assigned labels to the data, which were subsequently manually checked and modified as needed. This hybrid method greatly reduced labeling time without sacrificing accuracy. Combined with the public data, this resulted in around 14k image/label pairs for fine-tuning the classification model.

\begin{figure*}[tpb!]
    \centering
    \includegraphics[width=.72\linewidth]{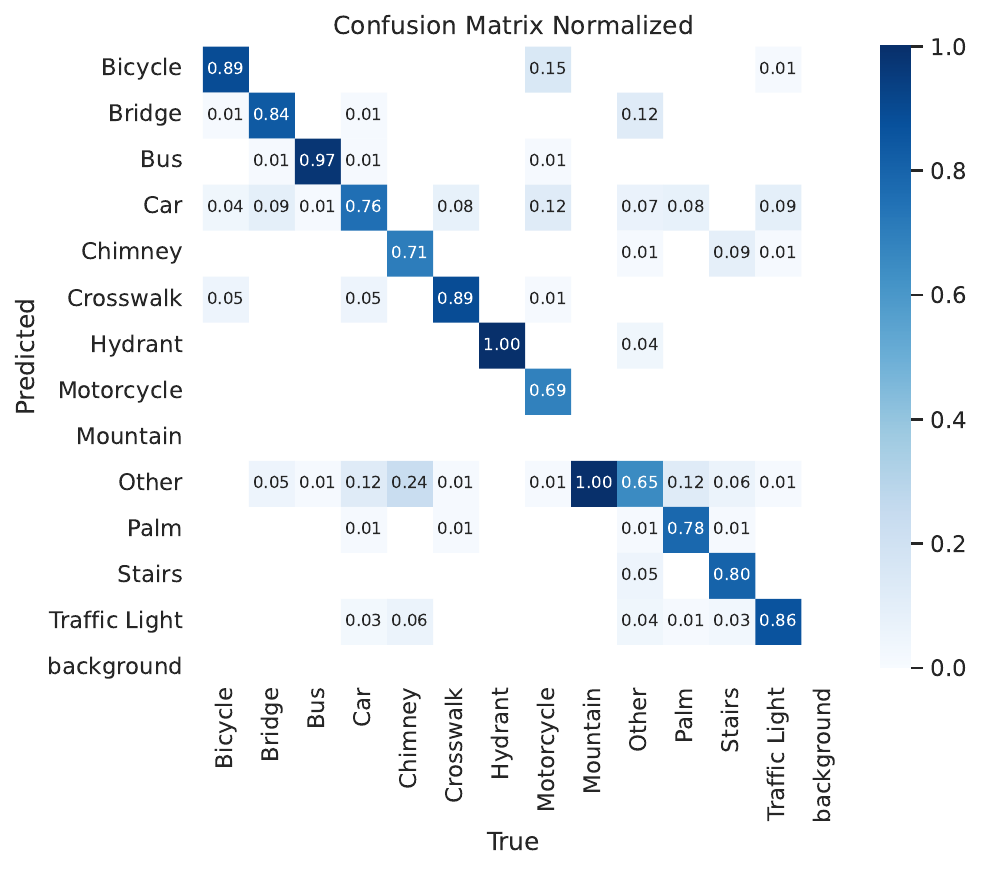}
    \caption{Normalized confusion matrix of the fine-tuned YOLOv8 model evaluated on the 13 classes seen in captcha challenges. The top 1 accuracy is 82.4\% while the top 5 accuracy is 99.5\%. The matrix highlights the model's ability to correctly classify various objects such as bicycles, bridges, buses, cars, and more, with values indicating the proportion of correct predictions. For example, bicycles were correctly identified with an accuracy of 89\%, while bridges and buses had an accuracy of 84\% and 97\%, respectively. The matrix reveals the strengths and weaknesses of the model in different classes, showing high precision in certain categories, such as hydrants (100\%), and notable confusion in others, such as varied performance in the identification of cars, illustrating the challenges of distinguishing between closely related objects.}
    \label{fig:confusion-matrix}
\end{figure*}

\paragraph{Segmentation}
When it comes to the segmentation task, type 2 represents a distinct challenge in contrast to classification. Precise identification of the individual segments that contain the target objects is necessary. Given the limited availability of publicly accessible datasets specifically designed for segmenting captcha images, and the extensive effort needed to gather and annotate such data, this study adopted an alternative methodology. We utilized the pre-trained YOLOv8 model specifically designed for segmentation tasks. This model was already trained on a broad set of classes, many of which are relevant to objects commonly found in \recaptcha{} image grids. The use of this pre-trained model allowed us to bypass the need for a large, labeled dataset specific to \recaptcha{} segmentation tasks.

\section{Proposed solution}
First, the captcha type and target class of the challenge are extracted. Second, based on the captcha type, the relevant setting of the YOLO v8 model is then used. For type 1 and type 3, the images in the grid are all classified, implying that the YOLO model predicts a class probability for each of the 13 classes, and if the target class has a probability of more than 0.2, then the model selects the image. For type 3 the process is repeated for all new images. To improve the model, we fine-tune the YOLO v8 classification model on the samples mentioned above in \cref{sec:Data}.

For type 2 the model segments the image and chooses any cell that overlaps with the segmentation. YOLO v8 for segmentation is only trained for 9 of the 13 possible classes, so if one of the 4 remaining classes appears, then the program will just skip the challenge. 

\section{Methodology}

\subsection{Evaluation environment}
The foundation of our experimental study is a customized testing environment, designed to allow a complete assessment of captcha solving strategies. The core of this environment is Python 3.9, which was chosen for its established reliability and the wide range of scientific computing tools it offers.

The Selenium WebDriver for Firefox is an essential part of our testing infrastructure, working alongside Python. This combination offers a highly accurate simulation of web browsing scenarios, which is essential to accurately present and interact with captchas in a way that closely mimics real-life user behavior. The automation features of Selenium, along with the reliability and efficiency of the Firefox browser, guarantee that every captcha instance is systematically handled and executed in consistent and replicable circumstances. The careful setup of Python 3.9 and Selenium with Firefox is a tactic to ensure that our methodology adheres to the highest levels of experimental precision. \footnote{We evaluate our system on Google \recaptcha{} demo site \url{https://www.google.com/recaptcha/api2/demo}.}

\subsection{Experiments}

\paragraph{VPN}
The integrity and authenticity of the test environment are enhanced by the implementation of a VPN connection, a crucial element in our experimental design. Our methodology acknowledges the advanced capabilities of captcha systems in identifying and reacting to multiple access attempts from a single IP address. To address this, we employ a VPN connection to dynamically change the IP address for each test run. This method guarantees that every interaction with the captcha challenges is seen as a distinct session, therefore reducing the possibility of being identified as suspicious by the security algorithms. Nevertheless, this strategy has its limitations. An evident drawback is the existence of additional network traffic on the VPN, which is a typical attribute of shared network services. The presence of unnecessary traffic can introduce fluctuations in network conditions, which can impact the assessment of the reCAPTCHA.

\paragraph{Mouse Movement}
An essential element in emulating human interactions with captchas involves replicating natural mouse movements. Our strategy utilizes Bézier curves, a mathematical concept frequently employed in computer graphics to represent smooth and scalable curves, to accurately simulate the movement of a user's mouse cursor. By utilizing Bézier curves, our approach accurately replicates the natural and unpredictable movement of a human cursor.

\paragraph{History and Cookies}
The last experiment will focus on the addition of history and cookies\footnote{Web pages should only have access to cookies, but we included all the browser data to ensure that an empty browser history could not impact the results.} to the browser session. For this, we use data from a real user to ensure that the data have a history and variability that would be seen from real users. 

\section{Results}
This section presents the results of the experiments. 

\subsection{VPN Usage Versus Non-VPN Conditions}

\begin{figure}[htbp]
    \centering
    \begin{subfigure}[c]{\columnwidth}
        \includegraphics[width=0.95\linewidth]{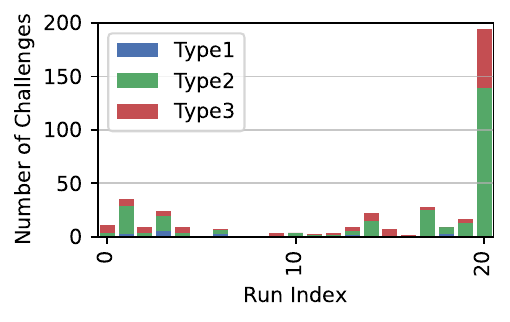}
        \caption{Number of solved challenges per captcha without VPN}
        \label{fig:without_vpn}
    \end{subfigure}%
    \hfill
    \begin{subfigure}[c]{\columnwidth}
        \includegraphics[width=0.95\linewidth]{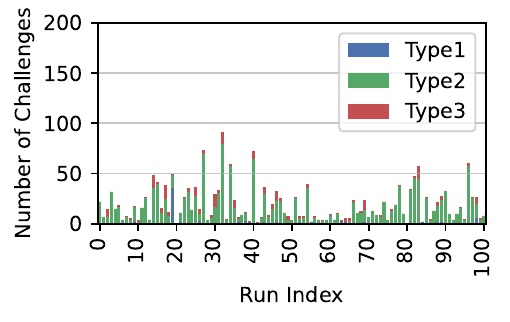}
        \caption{Number of solved challenges per captcha with VPN}
        \label{fig:with_vpn}
    \end{subfigure}
    \caption{Comparative analysis of captcha-solving challenges with and without the use of a VPN. The upper graph (a) shows the challenges without VPN, where the bot is flagged after the 19th run. The lower graph (b) demonstrates consistent performance over 100 runs with VPN, avoiding bot detection and subsequent challenge escalation.}
    \label{fig:vpn_comparison}
\end{figure}

\begin{table}[b!]
    \centering
    \begin{tabular}{lcc}
        \toprule
        & W/O VPN & With VPN  \\
        \midrule
        Minimum & 1     & 1     \\
        Median  & 9.00  & 13.00 \\
        Mean    & 19.10 & 19.23 \\
        Maximum & 194   & 91    \\
        Std.    & 40.20 & 17.54 \\
        IQR     & 13.00 & 20.00 \\
        \bottomrule
    \end{tabular}
    \caption{Statistical comparison of the number of necessary solved captcha challenges with and without a VPN connection. Std. is the standard deviation, and IQR is the interquartile range, the difference between the 75th and 25th percentile; both are used to assess the variability. The addition of the VPN connection improved the maximum, but the median and IQR are now much higher}
    \label{tab:stats_vpn}
\end{table}

The results of the experiment can be seen in \cref{fig:without_vpn} with the statistical data seen in \cref{tab:stats_vpn}.
The results emphasize the crucial need to use a VPN to reduce the likelihood of being identified as a bot by the risk assessment algorithms of \recaptcha{}. In the absence of a VPN, the bot faces an increasing number of difficulties beyond a specific limit of captchas. At first, the bot was capable of successfully solving captchas with a consistently low and steady number of challenges in each run. However, there was a significant change after the 20th run, as illustrated in \cref{fig:without_vpn}.

The graph shows the number of challenges required to successfully solve captchas in consecutive runs. During the initial 20 runs, challenges were tackled with a moderate and steady number of challenges. However, in the 21st iteration, there is a significant increase in the number of obstacles encountered, causing the bot's progress to come to a halt. The bot was unable to solve the captcha and the experiment was stopped after $\sim$200 challenges. This is also seen in the statistics, as the median is much lower without the VPN connection. 

The abrupt increase in the number of challenges indicates that the \recaptcha{} system most probably identified the bot's actions as suspicious, prompting a security mechanism that considerably increases the number of challenges. It demonstrates the system's ability to adjust to apparent automated behavior by implementing a defensive approach that gradually increases the number of challenges, eventually reaching a level that is almost impossible to overcome.

Therefore, the use of a virtual private network (VPN) is crucial in this particular situation. A VPN limits the ability of risk assessment algorithms to monitor and create a profile of the bot over several runs by allocating a different IP address for each run. This enables each captcha to be regarded as a distinct entity, bypassing the \recaptcha{} system's increasing security measures and hindering the bot from being obstructed or trapped in an infinite cycle of challenges. The bot using a VPN network could successfully pass the captcha for all 100 runs.

\subsection{Mouse Movement}
\begin{figure}[th]
\centering
\begin{subfigure}[b]{\columnwidth}
    \includegraphics[width=0.95\linewidth]{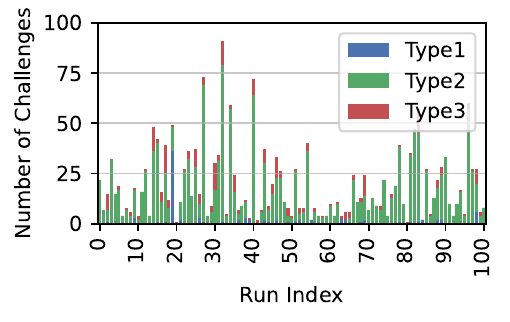}
    \caption{Without mouse movement. This is the same as \cref{fig:with_vpn} but the scale is different to make comparisons easier.}
    \label{fig:without_mouse}
\end{subfigure}
\hfill
\begin{subfigure}[b]{\columnwidth}
    \includegraphics[width=0.95\linewidth]{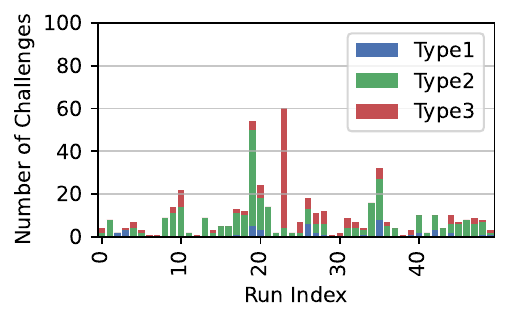}
    \caption{With straight line movement}
    \label{fig:with_straight_lines}
\end{subfigure}
\hfill
\begin{subfigure}[b]{\columnwidth}
    \includegraphics[width=0.95\linewidth]{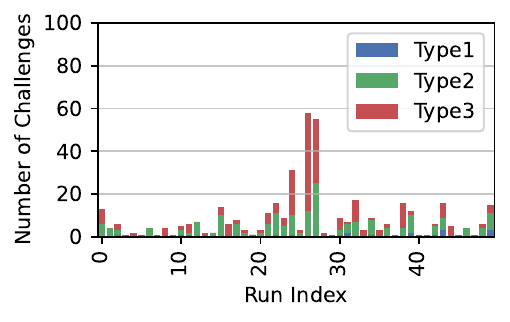}
    \caption{With Bézier curve movement}
    \label{fig:with_bezier_curves}
  \end{subfigure}
  \caption{Comparison of different mouse movement strategies in captcha solving. From top to bottom: (a) without mouse movement, (b) with straight line movement, and (c) with Bézier curve movement, illustrating the progressive performance improvement.}
  \label{fig:mouse_movement_comparison}
\end{figure}

In the previous tests, the mouse did not move while solving the challenges; therefore, we now include mouse movements. The incorporation of linear mouse motion into the captcha-solving algorithm has resulted in a noticeable enhancement in the bot's efficiency, as evidenced by the statistics in \cref{tab:stats_mouse}. \cref{fig:mouse_movement_comparison} depicts the number of challenges required to successfully solve a captcha during multiple iterations. This figure presents a comparison between the baseline method \cref{fig:without_mouse}, where the bot used the JavaScript click function of Selenium, and the improved strategy \cref{fig:with_straight_lines} that involved moving the mouse cursor directly to the target elements on the captcha grid.
\begin{table}[t!]
    \centering
    \begin{tabular}{lccc}
        \toprule
        & W/O mouse cursor & Straight Lines & Bézier Curves \\
        \midrule
        Minimum & 1     & 1     & 1     \\
        Median  & 13.00 & 7.00  & 5.00  \\
        Mean    & 19.23 & 9.72  & 8.38  \\
        Maximum & 91    & 60    & 58    \\
        Std.    & 17.54 & 11.56 & 11.45 \\
        IQR     & 20.00 & 7.75  & 7.00  \\
        \bottomrule
    \end{tabular}
    \caption{Statistical comparison of how many captcha challenges needed to be solved with and without using the mouse cursor, where the cursor movement was either in straight lines or along Bézier curves. Moving the mouse, regardless of the path type, improved (reduced) the overall number of challenges. However, a t-test comparing the effects of straight lines versus Bézier curves gave a t-statistic of 0.58 and a p-value of 0.57, indicating that there are no significant differences between the two movement types.}
    \label{tab:stats_mouse}
\end{table}

The findings demonstrate a significant decrease in the number of challenges needed to solve captchas for all types when using straight-line mouse movement. The incidence of encountering peak issues with each form of captcha, particularly type 3, has dropped considerably. This enhancement highlights the importance of using simulations that closely resemble human interactions. Conventional automated clicks can be easily detected by advanced bot detection systems used by captcha systems. However, incorporating linear mouse trajectories adds a more realistic imitation of human behavior, making it less likely that bots will be immediately identified and subjected to further challenges.

These findings indicate that the mouse path towards the target, rather than just the encounter with the target itself, is a crucial element in successfully passing the captchas. The graph presents empirical data that supports the concept that integrating mouse movements that resemble those of humans can greatly improve the efficiency of automated captcha-solving systems.

The integration of Bézier curve-based mouse movements into our captcha-solving algorithm has resulted in an additional enhancement in performance, surpassing the original increase achieved by integrating straight-line movements. The graph in this section illustrates the progress of the bot's capacity to overcome captchas by imitating the non-linear and smoother mouse movements facilitated by Bézier curves.

This advanced technique for moving the mouse greatly decreased the number of challenges needed to solve captchas of all kinds, especially the more challenging type 3 captchas. By including Bézier curves, the behavior of cursor navigation closely imitates the subtle actions of humans, making it more difficult for complex bot protection mechanisms like captcha systems to identify. Consequently, bot interactions have a reduced likelihood of triggering security defenses, which can result in an increased number of challenges or complete denial of access.

Additional statistical analysis was performed to compare the efficacy of mouse movements with straight lines and Bézier curves in solving captchas. A t-test was performed to evaluate the disparity in the total number of challenges needed to solve the captchas between these two methods. The analysis resulted in a t-statistic of 0.58 and a p-value of 0.57. This implies that although the use of Bézier curve movements showed a performance improvement, as evidenced by the decrease in the average number of challenges (8.38 for Bézier curves compared to 9.72 for straight lines, as seen in \cref{tab:stats_mouse}), the difference was not statistically significant.

\subsection{Impact of Browser History and Cookies on Captcha Solvability}

\begin{figure}[ht]
    \centering
    \begin{subfigure}[c]{\columnwidth}
        \centering
        \includegraphics[width=0.95\linewidth]{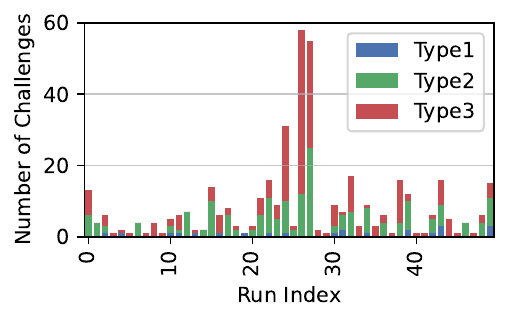}
        \caption{Without browser history and cookies. This is the same as \cref{fig:with_bezier_curves} but the scale is different to make comparisons easier.}
        \label{fig:without_cookies_and_history}
    \end{subfigure}%
    \hfill
    \begin{subfigure}[c]{\columnwidth}
        \centering
        \includegraphics[width=0.95\linewidth]{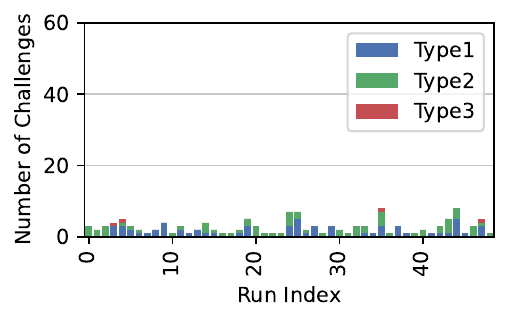}
        \caption{With browser history and cookies.}
        \label{fig:with_cookies_and_history}
    \end{subfigure}
    \caption{Comparative analysis of captcha-solving challenges with and without browser history and cookies from a browser session with a logged-in Google account. The left graph (a) displays the number of challenges in the absence of cookies and history, while the right graph (b) shows the number of challenges with cookies and history present, indicating the impact of user data on captcha challenge complexity.}
    \label{fig:history_cookies_comparison}
\end{figure}

\begin{table}[t]
    \centering
    \begin{tabular}{lcc}
        \toprule
        & \multicolumn{2}{c}{History and Cookies} \\
        & W/O & With \\
        \midrule
        Minimum & 1     & 1     \\
        Median  & 5.00  & 2.00  \\
        Mean    & 8.38  & 2.71  \\
        Maximum & 58    & 8     \\
        Std.    & 11.45 & 1.88  \\
        IQR     & 7.00  & 2.00  \\
        \bottomrule
    \end{tabular}
    \caption{Statistical comparison of the number of necessary solved captcha challenges when including or excluding cookies and browser history from a real-world active user. Including the data drastically reduces all statistics. In particular, the variability is much lower, which implies that the addition of browser data makes the performance more stable.}
    \label{tab:stats_history_and_cookies}
\end{table}

The results and statistics of the experiment can be seen in \cref{fig:history_cookies_comparison} and \cref{tab:stats_history_and_cookies}, respectively. The results show a drastic drop in the number of challenges required to solve a captcha. Additionally, a statistical test was performed. With a t-statistic of 3.42 and a p-value of 0.00, the tests show a substantial and statistically significant difference in \recaptcha{}-solving performance by including browser history and cookies. These findings indicate that the inclusion of browser history and cookies significantly affects the solvability of \recaptcha{}.

This discovery is significant and indicates the importance of user-specific data in \recaptcha{} challenges. \recaptcha{} seem to be less challenging to complete when there is browser history or cookies, probably because the security system recognizes an existing trusted user. On the other hand, if a user has no browsing history and cookies, the system may respond by presenting additional captchas, assuming that the user is less likely to be a genuine human.

The findings emphasize the flexible characteristics of contemporary captcha systems, which modify their level of difficulty according to user behavior and past experiences.

\subsection{Comparative Analysis of Human versus Bot Performance in Captcha Solving}

\begin{figure}[tbp]
    \centering
    \begin{subfigure}[c]{\columnwidth}
        \centering
        \includegraphics[width=0.95\linewidth]{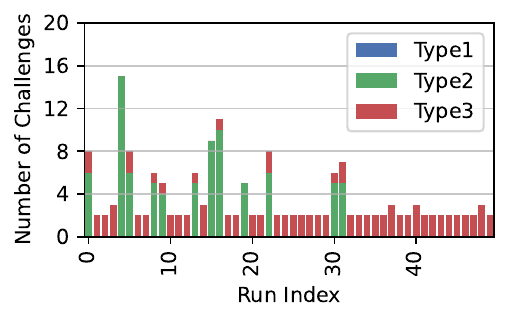}
        \caption{Number of challenges the human received to solve captchas.}
        \label{fig:human_challenges}
    \end{subfigure}%
    \hfill
    \begin{subfigure}[c]{\columnwidth}
        \centering
        \includegraphics[width=0.95\linewidth]{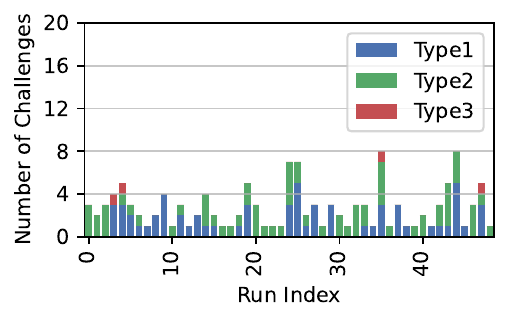}
        \caption{Number of challenges the bot received to solve captchas. This is the same as \cref{fig:with_cookies_and_history}, but the scale is different to make comparisons easier.}
        \label{fig:bot_challenges}
    \end{subfigure}
    \caption{Comparative performance of a human and a bot in solving captchas. The upper graph shows the number of challenges for the human user, while the lower graph shows the number of challenges for the bot. Both bot and human use about the same number of challenges per captcha; however, notably humans are mostly served type 3 captchas while our bot is mostly served types 1 and 2, and the human always has to solve at least two challenges.}
    \label{fig:human_vs_bot_comparison}
\end{figure}

This study performed a comparative analysis to assess human and bot performance in answering captchas under the same settings (with VPN, within the Selenium Browser, and with cookies and history). The experiment included a sequence of trials in which both a human subject and an automated bot attempted to solve the \recaptcha{} image-captchas. The results of the experiment can be seen in \cref{fig:human_vs_bot_comparison,tab:stats_human_vs_bot}.

The results of our study demonstrate that on average the bot exhibited a slightly lower number of challenges to successfully solve the captchas compared to the human solver, as illustrated in the table provided. However, a t-test was performed to assess the statistical significance of differences in efforts between the bot and the human. The resulting p-value was 0.11. The value, which exceeds the standard alpha level of 0.05, indicates that the observed variations in the number of challenges may be attributed to random chance and do not have statistical significance. This means that the bot has a similar behavior to the human statistics. However, as the p-value is relatively low, there is some evidence that there might be a difference between the bot and the human solver. Further studies could investigate this further.

\begin{table}[t!]
    \centering
    \begin{tabular}{lcc}
        \toprule
        & Bot & Human \\
        \midrule
        Minimum & 1     & 2     \\
        Median  & 2.00  & 2.00  \\
        Mean    & 2.71  & 3.50  \\
        Maximum & 8     & 15    \\
        Std.    & 1.88  & 2.79  \\
        IQR     & 2.00  & 1.00  \\
        \bottomrule
    \end{tabular}
    \caption{Statistical comparison of the number of necessary solved captcha challenges between our bot and a human captcha solver. We see that the bot overall has a lower mean and standard deviation than the human solver. However, when performing a t-test we get a t-statistic of 1.63 and a p-value of 0.11. So the difference is not significant at a level of 0.05 or 0.10, but there is some evidence of a potential difference between the two.}
    \label{tab:stats_human_vs_bot}
\end{table}

Comparison of human and automated bot performance in overcoming captchas provides detailed insights into the advancing capabilities of machine learning algorithms. The absence of a statistically significant disparity in the number of challenges between humans and bots, as evidenced by the t-test findings, questions the conventional belief that image-based captchas are an accurate way of distinguishing between human and nonhuman users. This discovery reinforces the need for ongoing improvement of captcha methods to stay up-to-date with the progressing field of artificial intelligence.

Moreover, the bot's success raises inquiries regarding the capacity of advanced bots to bypass security protocols that several online services depend on. It highlights a competition between creators of security systems and programmers of bots, where every improvement in bot technology leads to a matching upgrade in security measures.

Finally, we see that our solution is never blocked by the \recaptcha{} system, while \citet{wang2020using} and \citet{sivakorn2016m} only solved 68-71\% of the captchas. 

\section{Conclusion}
This study aims to evaluate the current status of image-based captcha challenges, with a specific focus on Google's \recaptcha{} and its vulnerability to advanced machine learning techniques. By conducting systematic experiments, we have shown that automated systems using advanced AI technologies, such as YOLO models, can successfully solve image-based captchas.

Comparative analysis of captcha-solving challenges by humans and bots demonstrated that although bots can closely mimic human performance, the observed difference was not statistically significant. This finding raises doubts about the reliability of image-based captchas as a definitive method for distinguishing between humans and bots. Our findings indicate that current captcha mechanisms are not immune to the rapidly advancing field of artificial intelligence. Additionally, we find that including browser cookies and history gives a substantial reduction in the number of challenges one is faced with. Our final model can solve 100\% of the presented captchas, while other models can only solve 68-71\% of the presented captchas from \recaptcha{}.

Continuous progress in AI requires a simultaneous development of digital security measures. Subsequent investigations should prioritize the development of captcha systems capable of adjusting to the complexity of artificial intelligence or explore alternative methods of human verification that can withstand the progress of technology.

Future studies might consider expanding the number of runs in every experiment. Currently, our study covers a range of 50 to 100 runs for every experimental configuration. Nevertheless, carrying out a larger quantity of iterations, possibly ranging in the hundreds or thousands, could yield more extensive observations about the enduring efficiency and dependability of the captcha-solving techniques. An expansion of this kind would provide a deeper understanding of the adaptive reactions of captcha systems over time and the enduring effectiveness of automated solving methods.

Future studies should improve the type 2 captcha dataset, which requires image segmentation. Some object classes from Google's \recaptcha{} are missing from our dataset, including the 'stairs' class. Future research should prioritize data collection to capture and label more objects to close this gap.
Furthermore, it would be beneficial for future research to investigate the threshold at which continuous captcha solving occurs before triggering a block. Due to the influence of cookies and user session data on captcha challenge difficulty, there is a valid risk that multiple attempts to solve captchas from the same computer with the same cookies could result in the computer being blocked by captcha systems. Conducting a thorough examination of the number of attempts required to activate countermeasures would provide valuable information.

The use of Google's \recaptcha{} has played a crucial role in improving website security on the Internet by successfully differentiating between actual users and automated bots. It fulfills various practical applications, tackling some of the most urgent security issues on the Internet. For example, \recaptcha{} addresses the scraping issue, which undermines the uniqueness of material by preventing automated theft to divert advertising income or gain a competitive advantage. This has become more relevant with the popularity of Large Language Models, LLMs, and the massive amounts of data required to train them \cite{brown2020language}. 
Our findings mark a crucial point in the ongoing dialogue between AI capabilities and digital security. They highlight the necessity for captcha technologies to evolve proactively, staying ahead of AI's rapid advancements. This is not just an academic challenge; it is a vital step toward ensuring the continued reliability and safety of our online environments.

\bibliographystyle{IEEEtranN}
\bibliography{main}

\end{document}